\documentclass[conference,letter]{IEEEtran}
\pdfoutput=1
\IEEEoverridecommandlockouts
\usepackage{amsmath,amssymb,amsfonts}
\usepackage{algorithmic}
\usepackage{graphicx}
\usepackage{textcomp}
\usepackage[T5]{fontenc}
\usepackage[utf8]{inputenc}
\usepackage{tabularx,ragged2e}
\usepackage{pifont}
\usepackage{float}
\usepackage{booktabs}
\usepackage{multirow}
\usepackage{forest}
\usepackage{tikz-qtree}
\usepackage{balance}
\usepackage{hyperref}
\usepackage{cleveref}
\usepackage{hyperref}

\hypersetup{colorlinks,allcolors=blue}
\DeclareMathOperator*{\argmax}{arg\,max}
\usepackage[backend=biber,
sorting=none,
doi=false,
isbn=false,
url=true,
maxcitenames=2,
mincitenames=1,
style=ieee,]{biblatex}

\makeatletter
\newcommand{\newlineauthors}{%
  \end{@IEEEauthorhalign}\hfill\mbox{}\par
  \mbox{}\hfill\begin{@IEEEauthorhalign}
}
\makeatother

\begin{document}

\title{An Empirical Study for Vietnamese Constituency Parsing with Pre-training}
\author{\IEEEauthorblockN{Tuan-Vi Tran}
\IEEEauthorblockA{University of Information Technology\\ Vietnam National University\\ Ho Chi Minh City, Vietnam\\
Email: 18520245@gm.uit.edu.vn}
\and
\IEEEauthorblockN{Xuan-Thien Pham}
\IEEEauthorblockA{University of Information Technology\\ Vietnam National University\\ Ho Chi Minh City, Vietnam\\
Email: 18520158@gm.uit.edu.vn}
\newlineauthors
\IEEEauthorblockN{Duc-Vu Nguyen}
\IEEEauthorblockA{University of Information Technology\\ Vietnam National University\\ Ho Chi Minh City, Vietnam\\
Email: vund@uit.edu.vn}
\and
\IEEEauthorblockN{Kiet Van Nguyen}
\IEEEauthorblockA{University of Information Technology\\ Vietnam National University\\ Ho Chi Minh City, Vietnam\\
Email: kietnv@uit.edu.vn}
\and
\IEEEauthorblockN{Ngan Luu-Thuy Nguyen}
\IEEEauthorblockA{University of Information Technology\\ Vietnam National University\\ Ho Chi Minh City, Vietnam\\
Email: ngannlt@uit.edu.vn}
}

\maketitle
\begin{abstract}
Constituency parsing is an important problem that gets more attention in natural language processing.
In this work, we use a span-based approach for Vietnamese constituency parsing. Our method follows the self-attention encoder architecture and a chart decoder using a CKY-style inference algorithm.
We present analyses of the experiment results of the comparison of our empirical method using pre-training models XLM-Roberta and PhoBERT on both Vietnamese datasets VietTreebank and NIIVTB1.
The results show that our model with XLM-Roberta archived the significantly F1-score better than other pre-training models, VietTreebank at 81.19\% and NIIVTB1 at 85.70\%. 
\end{abstract}

\section{Introduction}
The field of natural language processing (NLP), also known as
computational linguistics, which involves the engineering of computational models and processes to solve practical issues in understanding human languages that have leveraged the effect of modern proposed \cite{survey}. 
Particularly, constituency parsing has a variety of applications in NLP tasks, such as grammatical error correction \cite{dahlmeier-ng-2012-better}, question answering \cite{10.1145/1571941.1571975}, dynamic oracle \cite{fried-klein-2018-policy}, information extraction \cite{jiang-diesner-2019-constituency,zhang-etal-2006-exploring}.
\newline
Vietnamese constituency parsing has not been achieved a state-of-art performance as English, Chinese, etc. One main reason is the ambiguities in parsing typical sentences \cite{quy-nguyen-2018}. Another critical reason is the lack of massive and quality corpus for this task.
\newline
Current deep learning architectures such as recurrent neural networks (RNNs), in particular, Long Short-term Memory (LSTM) are widely applied to most NLP tasks with great success due to their ability to capture long-distance linguistic features without the need for feature engineering \cite{VN-span-based}. 
However, RNNs are not the best performance architecture in sequence-to-sequence (seq2seq) tasks, particularly constituency parsing in the present day. 
A new mechanism can be extended further into a seq2seq context to eventually replace the role of RNNs, which called attention mechanism.
\begin{figure}[ht]
\includegraphics[width=\columnwidth]{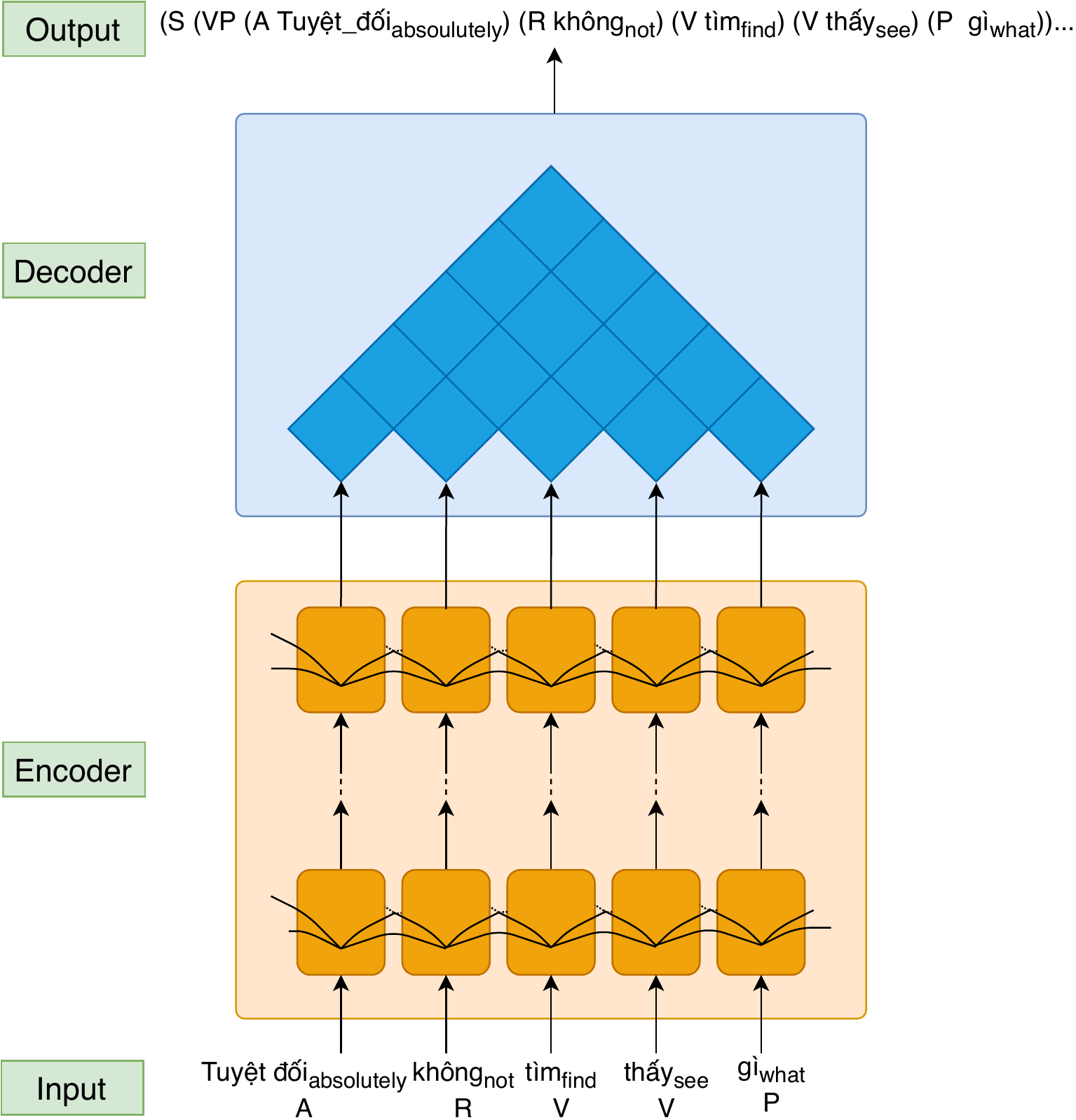}
\centering
\caption{Parser combines a chart decoder
with a sentence encoder based on self-attention}
\label{fig:figure1}
\end{figure}
\newline
In the paper of \cite{Kitaev-2018-SelfAttentive}, they proposed a state-of-the-art approach, a combination of an encoder using self-attentive architecture and a decoder adapted to parsing (see a Vietnamese example in Figure~\ref{fig:figure1}). In section 2, we re-describe self-attentive architecture. A mechanism handle with ambiguous problems, gather the information between words in the sentence. Each location can attend to other locations based on their positions or their contexts.
\newline
In section 3, we introduce both constituency parsing datasets: VietTreebank \cite{VLSP} and NIIVTB1 \cite{quy-nguyen-2018}, which are used for this work. We represent the general processing and make statistical the constituent tags. And in section 4, we proceed to evaluate and analyze experimental results after practicing self-attention architecture with a total of two groups pre-training models: $\text{XLM-R}_{\text{base}}$, $\text{XLM-R}_{\text{large}}$ \cite{conneau2019unsupervised} and $\text{PhoBERT}_{\text{base}}$, $\text{PhoBERT}_{\text{large}}$ \cite{phobert}, on both mentioned datasets. We compare the performance of these pre-training models and have conclusions. To sum up, our research makes the following contributions:
\begin{enumerate}
  \item The state-of-the-art performance system on constituency parsing for Vietnamese corpus.
  \item The comparison results between two pre-traning models: XLM-Roberta and PhoBert on the same datasets. 
\end{enumerate}

\section{Model}
In this work, we follow self-attention encoder architecture proposed by \cite{Kitaev-2018-SelfAttentive} (see in Figure~\ref{fig:figure1}). In order to get more information to analysis and evaluate exactly, we use four pre-training models: $\text{XLM-R}_{\text{base}}$, $\text{XLM-R}_{\text{large}}$ \cite{conneau2019unsupervised} and $\text{PhoBERT}_{\text{base}}$, $\text{PhoBERT}_{\text{large}}$ \cite{phobert}. We add a part-of-speech tagging component and auxiliary loss to the parser and do not use tags to the train/dev sets are seen as predicted tags.
\subsection{Tree Scores and Chart Decoder}
Each constituency parsing tree $T$, has score $S(T)$, which is calculated by sum of all the scores of internal nodes of label span trees:
\begin{IEEEeqnarray}{rCl}
S(T) = \sum_{(i,j,l)\in T } s(i,j,l)
\end{IEEEeqnarray}
Here, $i$ and $j$ are two location in a sentence, the span tree from $i$ to $j$ has label $l$. To handle with dummy node $\varnothing$ problem, we assign $\forall i, j : s(i, j, \varnothing) = 0$. To optimize model during test time, we use CKY-style algorithm \cite{gaddy} to find best score tree $\hat{T}$. The model is trained in order to with the gold parsing tree $T^{*}$ in such a way that for all tree $T$ satisfied the margin constraints:
\begin{IEEEeqnarray}{rCl}
S(T^{*}) \geq S(T) + \Delta (T,T^{*})
\end{IEEEeqnarray}
for all trees T, using the hinge loss function to minimize
\begin{multline}\label{eq6}
L_{c}=\max (0,\max\limits_{T\neq T^{*}}(S(T) + \Delta (T,T^{*}))-S(T^{*}))
\end{multline}
where $\Delta$ is Hamming loss on labeled spans. $L_{c}$ is equal to 0 when all constraint conditions are satisfied or the magnitude of the largest margin violation otherwise. Our final parsing goal is to find the highest-scoring valid tree is 
\begin{IEEEeqnarray}{rCl}
\hat{T} = \argmax_{T} S(T)
\end{IEEEeqnarray}

\subsection{Context-Aware Word Representations}

The encoder takes the list of vector embedding $[w_{1},w_{2},...,w_{n}]$ as input, head and tail of this list are embedded by $<\textit{START}>$ and $<\textit{STOP}>$ tokens. Moreover, the encoder contains a matrix positional embedding,with every number i={1,2,3,...} (up to some maximum the sentence length) is a vector $p_{i}$.
The encoder is composed of a stack N=8 identical layers. Each layer has two sub-layers namely multi-head self-attention layer mechanism and position-wise feed-forward sublayer. The output of each layer is $LayerNorm(x+Sublayer(x))$ with input x, here $Sublayer(x)$ is implemented by the sub-layer itself. We employ each sublayer by a residual connection and a Layer Normalization \cite{ba2016layer} step.
Result is outputs of dimension $d_{model}$.
\subsubsection{Self-Attention}
Self-attention is a form of attention mechanism when the model receives each word (of input sentence). Self-attention allows it to gather information from other words in the input sequence that can help lead to a better encoding for this word. The first sublayer consists of 8 layers is a multi-headed self-attention mechanism. Attention mechanism takes a $T \times d_{model} $ matrix $X$ as input, each word $t$ in the sentence is represented by row of vector $x_{t}$ respectively. 
\begin{figure}[ht]
\includegraphics[width=0.80\columnwidth]{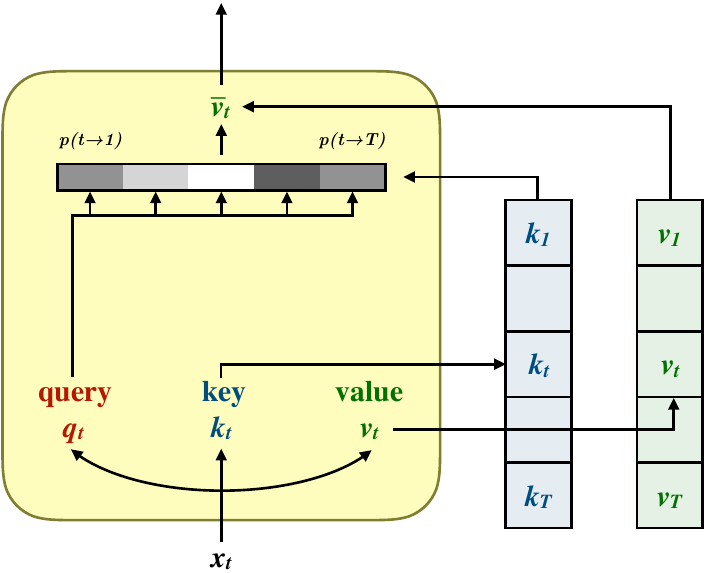}
\centering
\caption{A single attention head \cite{Kitaev-2018-SelfAttentive} }
\label{fig:figure2}
\end{figure}
\newline
Look at the Figure~\ref{fig:figure2} in more detail, a single attention head takes an input $x_{t}$, the first task of it is create three vectors: a query $q_{t}$, a key $k_{t}$, and a value $v_{t}$. These vectors are created by multiplying three weight matrices $W_{Q} \in \mathbb{R}^{d \times d_{k}},W_{K}\in \mathbb{R}^{d \times d_{k}},W_{V}\in \mathbb{R}^{d \times d_{v}}$ by the input $x_{t}$: a query $q_{t}=W_{Q}^\intercal x_{t}$, a key $q_{k}=W_{K}^\intercal x_{t}$, a value $q_{v}=W_{V}^\intercal x_{t}$, where number of dimensions of query vector and key vector have the same $d_{k}$, and number of dimensions of value vector are $d_{v}$. 
We proceed to compute the dot-product the query with all the keys, then divide by $d_{k}$ and use soft-max function to get the weight assigned to each value. 
\begin{multline}\label{eq7}
SingleHead(X) = \left[softmax(\frac{QK^\intercal}{\sqrt{d_{k}}})V \right]W_{O}
\end{multline}
where $W_{O} \in \mathbb{R}^{d_{v} \times d} $ is the output of the weight matrix, and with $L$ is the segment length of input sequence: $Q = XW_{Q} \in \mathbb{R}^{L \times d_{k}} $; $K = XW_{K} \in \mathbb{R}^{L \times d_{k}} $; $V = XW_{V} \in \mathbb{R}^{L \times d_{v}} $ are the weight matrix of query, key and value respectively. 
\newline
Rather than only computing the attention once, it is beneficial to linearly project the queries, keys and values h times with different. Multi-headed attention is presented by concatenating these single-head attentions.
\begin{multline}\label{eq8}
MultiHead(X) = \sum_{i=1}^{h} SingleHead^{(i)}
\end{multline}
where the projections are the trainable of parameters of each 8 heads, $W_{Q}^{(i)},W_{K}^{(i)} \in \mathbb{R}^{d\times d_{k}/h}$, $ W_{V}^{(i)} \in \mathbb{R}^{d\times d_{v}/h}$ are weight matrices to map input embedding of size $L\times d$ into query, key and value matrices; $W_{O}^{(i)} \in \mathbb{R}^{d_{v} \times d}$ is the output linear transformation. In this practice, we initial $h=8$ parallel attention heads. This allows a word to attend to information from up to 8 different positions in the sentence at each attention sublayer.
Another essential component of the transform is the Position-Wise Feed-Forward Sublayer (FFN). Follow the form of \cite{attention-allneed}. These sublayers are fully-connected layers with the dimensionality of input and output are the same.
\begin{multline}\label{eq9}
FeedForward = W_{2}relu(W_{1}x+b_{1})+b_{2}
\end{multline}

\subsection{Span score}
A span starts at $i^{th}$ position and ends at $j^{th}$ correspond with a span vector $v_{ij}=[\overrightarrow{\rm y_{j}}-\overrightarrow{\rm y_{i}};\overrightarrow{\rm y}_{j+1}-\overrightarrow{\rm y}_{i+1}]$. Span score $s(i,j,\cdot)$ function was implemented by proposed of \cite{stern2017minimal}. Putting the span representation into a one-layer feed-forward network whose output dimensionality equals the number of possible labels.

\begin{IEEEeqnarray}{rCl}
s(i,j,l)=[M_{2}relu(LN(M_{1}v_{ij}+c_{1}))]_{l}
\end{IEEEeqnarray}
where $LN$ is Layer Normalization, $relu$ is the Rectified Linear Unit non-linearity and $v$ is combination of all span score vectors in the sentences.
\section{Experiments}
\subsection{Datasets}
First, we describe the general of datasets. As mentioned, we use two well-known datasets to conduct experiments: VietTreebank \cite{VLSP09} and NIIVTB1\cite{quy-nguyen-2018} with bracketed structure format (similar to English Penn Treebank). 

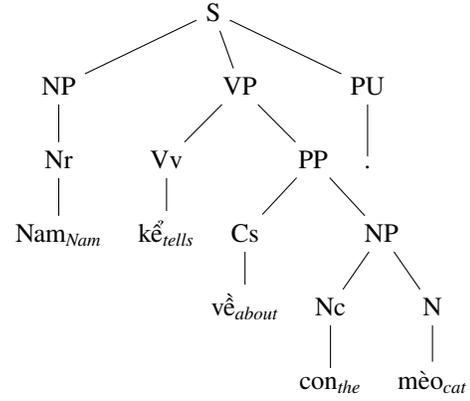
\begin{figure}[ht]
\centering
\begin{forest}
[S
    [NP [Nr [$\text{Nam}_{\textit{Nam}}$]]]
    [VP 
        [Vv [$\text{kể}_{\textit{tells}}$]]
        [PP [Cs [$\text{về}_{\textit{about}}$]]
            [NP
                [Nc [$\text{con}_{\textit{the}}$]]
                [N [$\text{mèo}_{\textit{cat}}$]]
            ]
        ]
    ]
    [PU [.]]
]
\end{forest}
    \medskip
    \caption{Illustrate a Vietnamese parse tree}
    \label{fig:figure3}
\end{figure}

In particular, the raw text of VietTreebank and NIIVTB1 were gathered from Youth (Tuoi Tre) newspaper\footnote{https://tuoitre.vn} cover social and political topics primarily. VietTreebank corpus totals only 16 constituent tags (see Figure 5 in \cite{VLSP}), while the constituent tags of NIIVTB1 were modified and have 18 types (see in table 14 \cite{quy-nguyen-2018}). Below (Figure~\ref{fig:figure3}) is example of illustrate from bracketing form sentence. Original sentence is ``Nam kể về con mèo'' (Nam tells about the cat).
\newline
These ambiguous problems in Vietnamese are the adjectives that can stand in front of or behind the nouns and the verbs in the noun phrases and the verb phrases, and the adjunct expresses the words without tense, number, context, ... \cite{nguyen-etal-2016-challenges} proposed the solution in this cases.

\begin{figure}[!ht]
\includegraphics[width=\columnwidth]{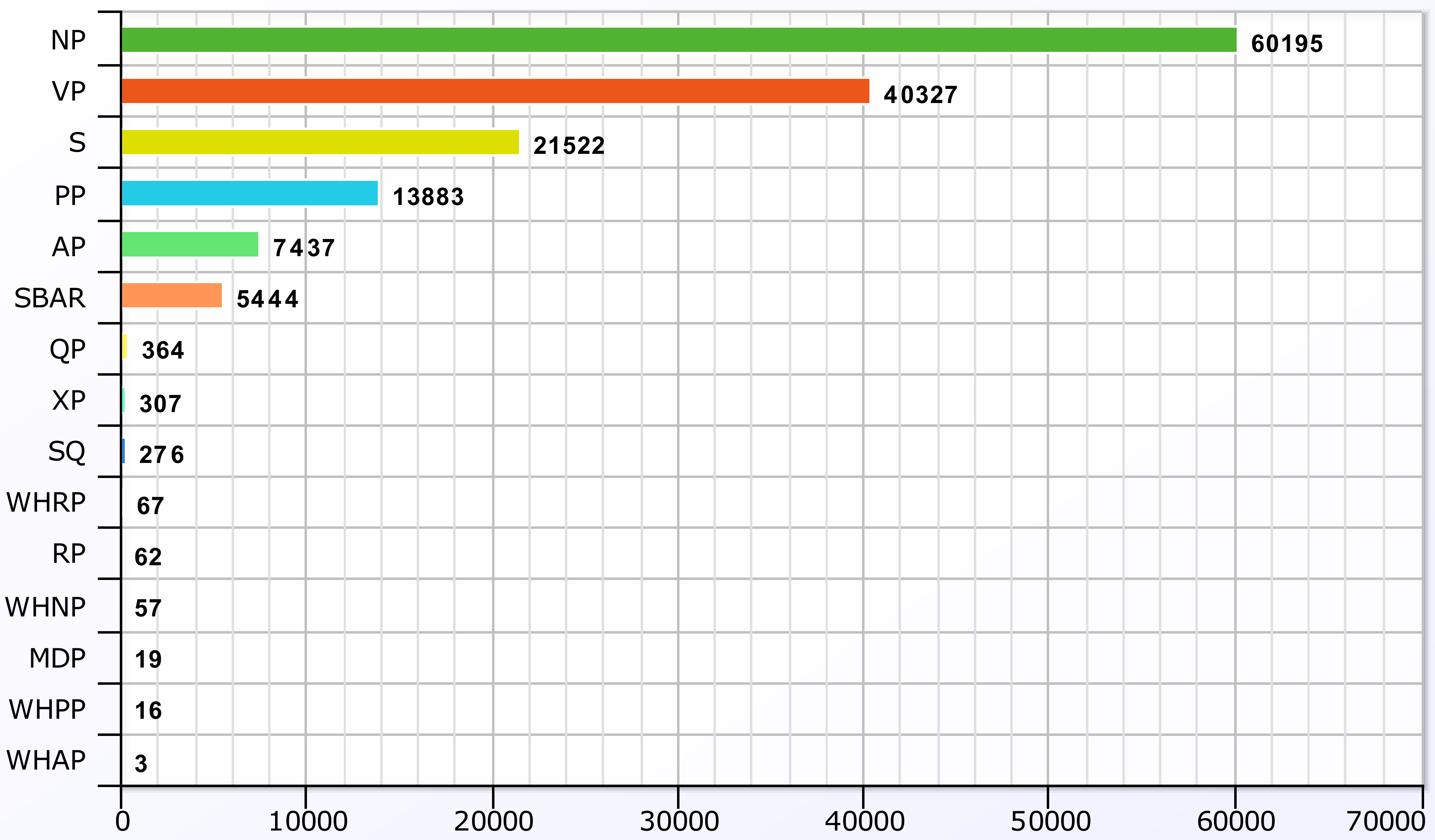}
\centering
\caption{Illustrate the statistics of constituent tags of VietTreebank dataset}
\label{fig:figure4}
\end{figure}

\begin{figure}[!ht]
\includegraphics[width=\columnwidth]{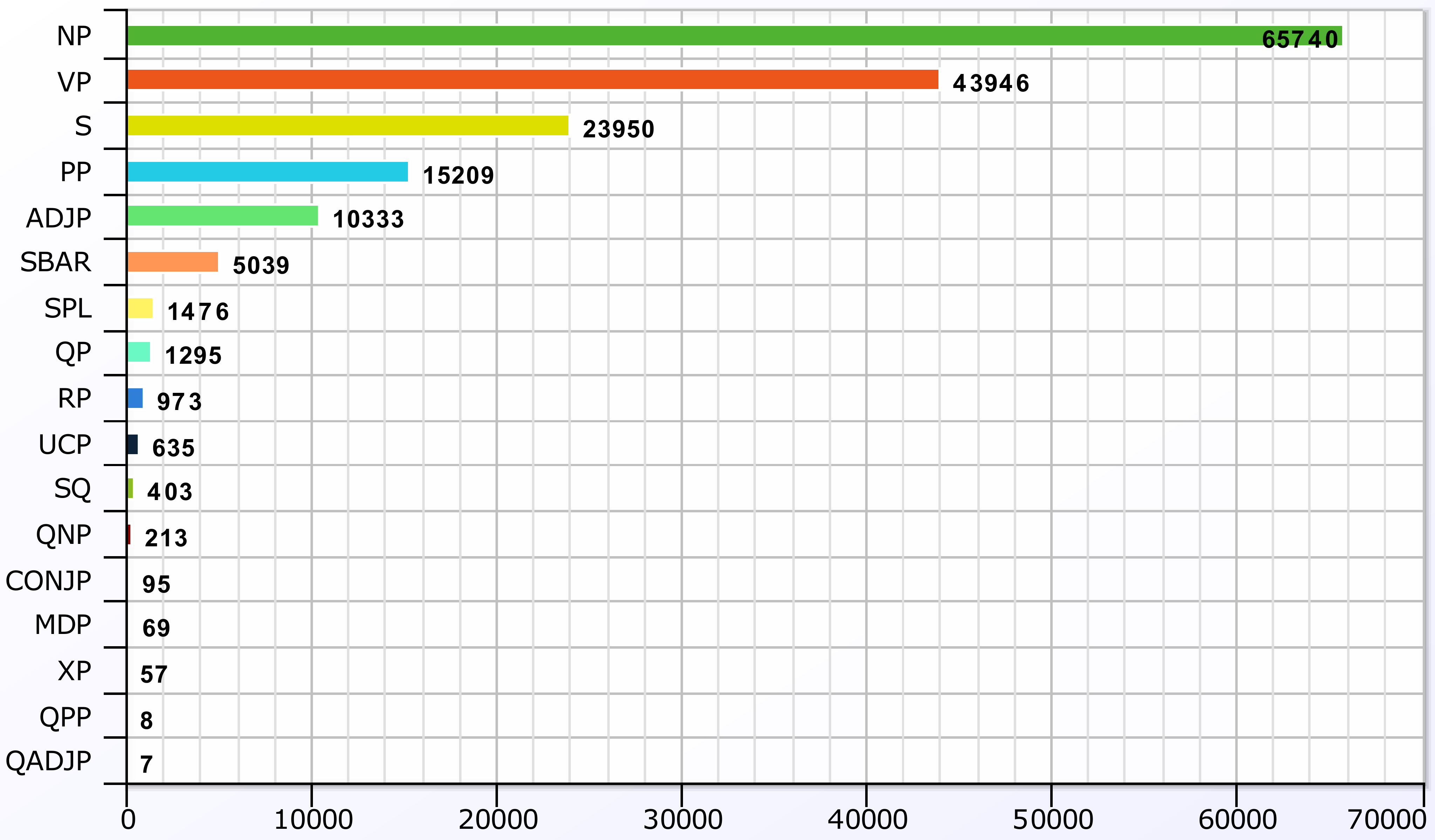}
\centering
\caption{Illustrate the statistics of constituent tags of NIIVTB1 dataset}
\label{fig:figure5}
\end{figure}

The distributions of labels appear in VietTreebank (Figure~\ref{fig:figure4}) and NIIVTB1 (Figure~\ref{fig:figure5}). The number of labels in both corpora is not the same. These figures show that the corpora are extraordinarily unbalanced and biased toward the popular constituent tags. We split train/test/dev sets follow  \cite{quy-nguyen-2018} for NIIVTB1. Because, the size of dev set of VietTreebank in \cite{Nguyen2014NLDB} is too small, we cut a part of train tail to dev to get the ratio train/dev is 8636:510 and preserve the test set.
\newline
\subsection{Measures}
Following the paper of \cite{Kitaev-2018-SelfAttentive}, we also use Precision ($P$), Recall
($R$), F1-score ($F_{1}$) to evaluate our models.
\begin{IEEEeqnarray}{rCl}
P=\frac{|T_{C}|}{|T_{P}|}; R = \frac{|T_{C}|}{|T_{G}|}; F_{1} = \frac{2 \times P \times R}{R + P}
\end{IEEEeqnarray}
where $|T_{C}|$ is number of correct constituent tags; $|T_{G}|$ is number of constituent tags in the gold tree; $|T_{P}|$ is number of predict constituent tags.
\subsection{Implement}
In this part, we present a summary of the initiation of our models. We experiment step-by-step with two pre-training models: XLM-Roberta\footnote{https://github.com/pytorch/fairseq/tree/master/examples/xlmr} and PhoBERT\footnote{https://github.com/VinAIResearch/PhoBERT} with self-attention architecture\footnote{https://github.com/nikitakit/self-attentive-parser}. For training stages, we set up main hyper-parameters with 2 layers of self-attentions, 150 max epochs, batch size is 150 and 1500 sub-batch max tokens.
\section{Analyses}
\subsection{Result}
The Table \ref{tab:table1} compares the constituency parsing obtained by the six parsers. The first two rows describe the performance mentioned in \cite{quy-nguyen-2018} and \cite{VN-span-based} in the same datasets. While the last four rows report the score of our systems.
\newline
As expected, the self-attention encoder architecture with XLM-R perform significantlt better than the previous parsers. Specifically, we witness 1.64\% and 9.75\% absolute improvements in VietTreebank and NIIVTB1 respectively.

\subsection{XLM-R base and PhoBERT base}

\begin{table*}[htb]
\caption{\label{tab:table1}The result of both test datasets on PhoBERT and XLM-R}
\begin{tabular}{|c|c|c|c|c|c|c|}
\hline
\multirow{2}{*}{Model} & \multicolumn{3}{c|}{\textbf{VietTreebank}} & \multicolumn{3}{c|}{\textbf{NIIVTB1}} \\ \cline{2-7} 
                       & P(\%)   & R(\%)  & F1(\%)          & P(\%)    & R(\%)    & F1(\%)           \\ \hline
Berkeley parser \cite{quy-nguyen-2018}                & 73.07   & 70.40  & 71.71           & 77.21    & 74.71    & 75.95            \\ \hline
BERT MINIMAL PARSER \cite{VN-span-based}               & 80.05   & 79.05  & 79.55  & -    & -    & -   \\ \hline
$\text{PhoBERT}_{\text{base}}$               & 81.14   & 79.60  & \textbf{80.36}  & 85.41    & 84.18    & \textbf{84.79}   \\ \hline
$\text{XLM-R}_{\text{base}}$                   & 79.95   & 78.61  & 79.28           & 83.59    & 82.60    & 83.09            \\ \hline
$\text{PhoBERT}_{\text{large}}$                & 80.55   & 80.54  & 80.55           & 85.32    & 84.50    & 84.91            \\ \hline
$\text{XLM-R}_{\text{large}}$                    & 80.78   & 81.61  & \textbf{81.19}  & 85.91    & 85.50    & \textbf{85.70}   \\ \hline
\end{tabular}
\centering

\end{table*}
As can see in Table~\ref{tab:table1}, the performance of PhoBERT on test sets is better than that of XLM-R. In the VietTreebank, while the F1-score of XLM-R could not reach at 80.0 \%, PhoBERT has a significant improvement (80.36\%). It has a similar previous in the NIIVTB1, F1-score of PhoBERT and XLM-R are 84.79\% and 83.09\% respectively. 

\subsection{XLM-R large and PhoBERT large}

Opposite the trend in section 4.2, there is a domination of XLM-R in large models group. The F1-score in the VietTreebank of XLM-R archived 81.19\%, a slightly increase of 0.64\%. Beside, XLM-R reached a peak at 85.70\% in NIIVTB1 test set whereas the result of PhoBERT is only 84.91\%. \newline
The next section gives a detailed accuracy and error analysis on constituent tags with two highest parser,$\text{PhoBERT}_{\text{large}}$ and $\text{XLM-R}_{\text{large}}$.
\subsection{Discussion}
We start to analysis the effect of the imbalance of VietTreebank dataset to both models. It is more precise when combining the Figure~\ref{fig:figure4} and Figure~\ref{fig:figure7}. The four label (S, NP, VP, PP)  took up the the most percentage in VietTreebank. It lead to $F_{1}$ score of them in two systems are over 80\%. (WHPP, MDP and WHAP) are three significant constituent tags reach almost 0\% in $F_{1}$ score. The rest of labels (except SQ) have performance archived from 45\% to over 65\% and the disparity between two systems is not large. The case in which SQ, there is a huge different of two $F_{1}$ score. XLM-R work with question sentences better than PhoBERT.
\begin{figure}[!ht]
\includegraphics[width=\columnwidth]{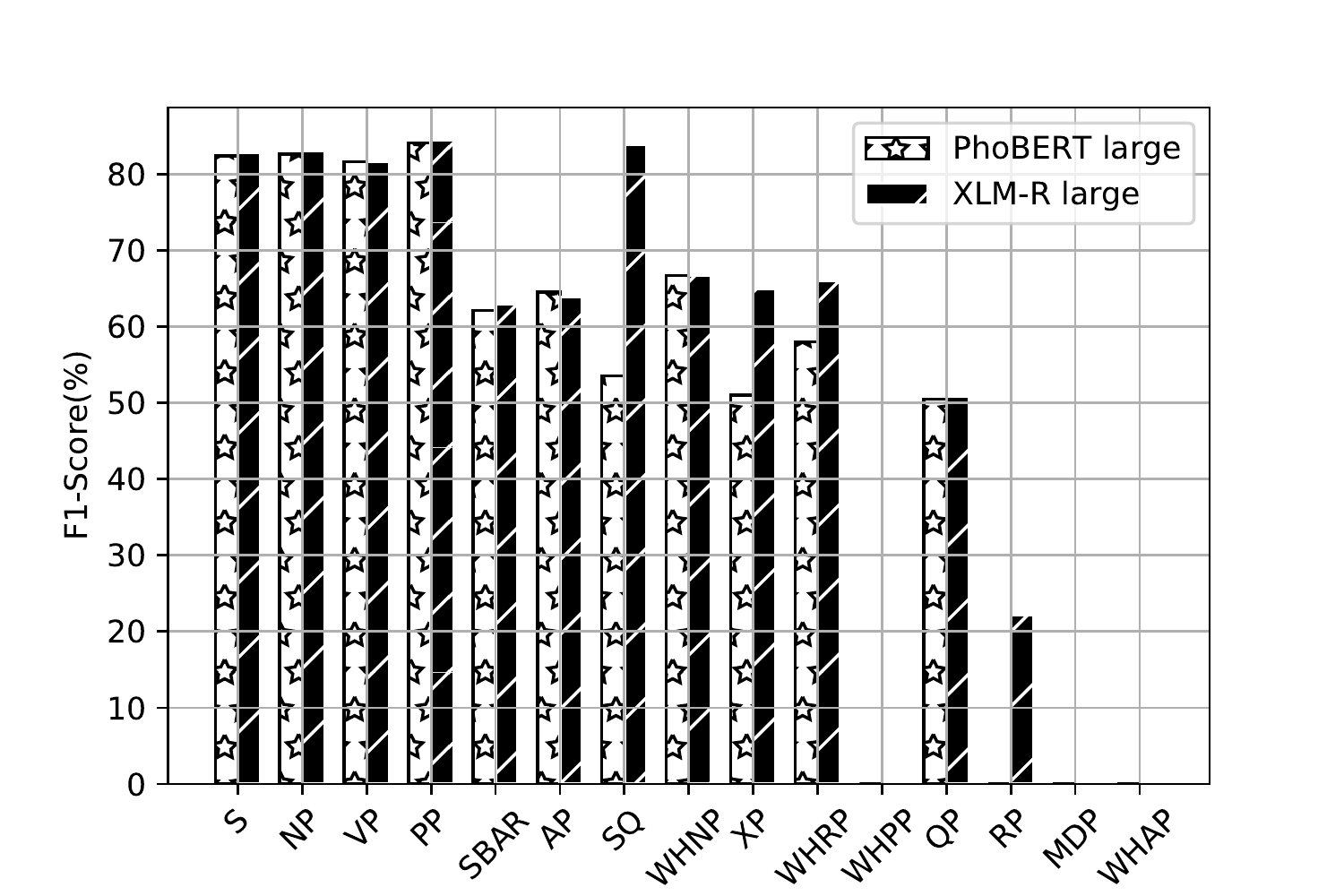}
\centering
\caption{F1-score of $\text{PhoBERT}_{\text{large}}$ and $\text{XLM-R}_{\text{large}}$ for each constituent tags on VietTreebank test set}
\label{fig:figure6}
\end{figure}
\begin{figure}[!ht]
\includegraphics[width=\columnwidth]{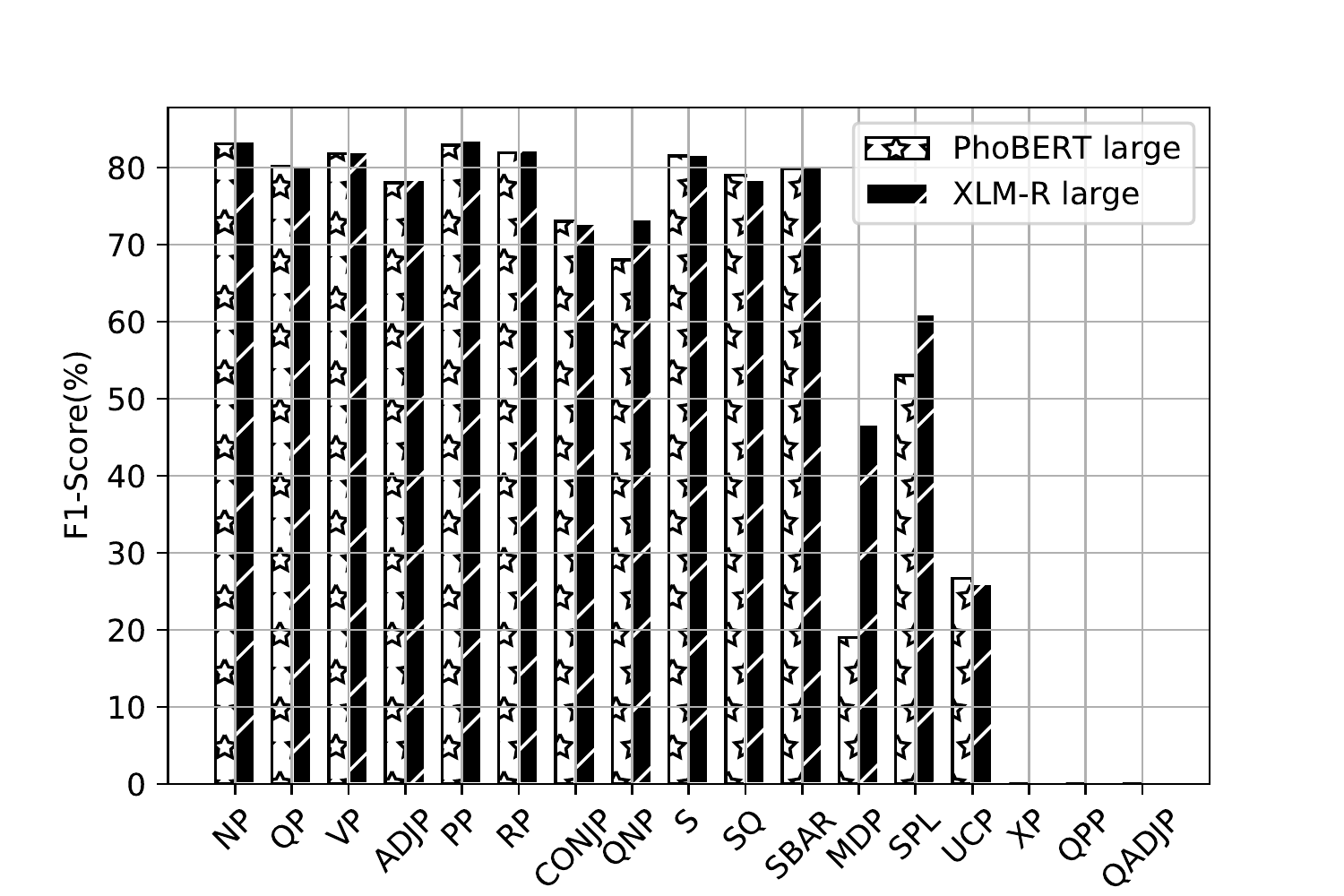}
\centering
\caption{F1-score of $\text{PhoBERT}_{\text{large}}$ and $\text{XLM-R}_{\text{large}}$ for each constituent tags on NIIVTB1 test set}
\label{fig:figure7}
\end{figure}
Looking in Figure~\ref{fig:figure7} for more detail about $F_{1}$ score on NIIVTB1 test set. (QPP, QADJP and XP) are three constituent tags made up the lowest proportion (see figure 5), thereforce, $F_{1}$ score of them are nearly 0\%. Almost there is a bit different between two models for each labels in this treebank. The group of four labels (NP, VP, PP and S) reach the best result (over 84\%) among 17 constituent tags. 

\section{Related Work}
In recent years, attention has become a popular and powerful mechanism in deep learning community.
It brings a strong development in not only NLP, but also a patch of domains such as recommendation \cite{attn-recommendation}, image processing \cite{Wang_2018}, speech recognition \cite{chorowski2015attentionbased}. 
\newline
The first version of attention was proposed is sequence to sequence (Seq2Seq) architecture for machine translation \cite{cho-etal-2014-learning,sutskever2014sequence}. They independently exported similar architecture including two RNNs namely encoder linking to decoder. 
However, there is an issue in seq2seq architecture that might lose some information when processing long sentences. 
Therefore, Align \& Translate \cite{LuongPM15,bahdanau2014neural} born to exploit the context vector to align the source and target. 
\newline
In year 2006, hierarchical attention network (HAN) \cite{yang-etal-2016-hierarchical} that attention can be effectively used on two levels: word and sentence level. Thereby it  allows the system to pay less or more attention to individual word and sentence accordingly when assembling the representation of a document.
\newline
Transformer neural network architecture \cite{attention-allneed} is one of the revolutionized studies in NLP field. It based entirely on attention mechanism, multi-headed self-attention is proposed instead of encoder-decoder architecture .The intuition behind Transformer got a lot of attentions, pave the way for the development of self-attention-based models such as the Bidirectional Encoder Representations from Transformers (BERT) archived novel result in many NLP tasks. A group of BERT-based models demonstrate high performances, such as XLNET \cite{XLNet}, RoBERTa \cite{conneau2019unsupervised}, DistilBERT \cite{sanh2019distilbert}, ALBERT \cite{lan2019albert} and PhoBERT \cite{phobert} have been proposed recently.

\section{Conclusion}
Despite the essential roles in providing valuable resources to a lot of NLP tasks, unexpectedly Vietnamese constituency parsing has not archived the good performance as English. 
This research contributes to extensive examinations of the Span-based approach for constituency Parsing of the Vietnamese language.
We report the highest performance up to date for Vietnamese consituency parsing with VietTreebank at 81.19\% and NIIVTB1 at 85.70\%. 
By using the pre-training models, our system archived state-of-the-art performance of parsing on both datasets (VietTreebank and NIIVTB1).
The experimental results illustrate the improvement of parsing accuracy which issues existed in the before studies on the same treebank.
\printbibliography

\end{document}